\pdfoutput=1

\documentclass[a4paper,11pt]{article}

\usepackage[preprint]{acl}

\usepackage[T1]{fontenc}
\usepackage[utf8]{inputenc}

\usepackage{times}

\usepackage{amsmath}
\usepackage{amsthm}
\usepackage{amssymb}

\usepackage{latexsym}

\usepackage{microtype}
\usepackage{inconsolata}

\usepackage{subfiles}

\usepackage[most]{tcolorbox}
\usepackage{booktabs}
\usepackage{multirow}
\usepackage{enumitem}
\usepackage{graphicx}
\usepackage{xcolor}
\usepackage{hyperref}
\usepackage[capitalise]{cleveref}
\usepackage{caption}

\usepackage{siunitx} 
\usepackage{CJKutf8}    
\usepackage[safe]{tipa} 
\usepackage{comment}
\usepackage{minted}     
\usepackage{csquotes}   

\renewcommand{\vec}[1]{\mathbf{#1}} 
\DeclareMathOperator*{\argmax}{arg\,max} 
\DeclareMathOperator*{\argmin}{arg\,min} 

\newcommand\blfootnote[1]{%
    \begingroup
    \renewcommand\thefootnote{}\footnote{#1}%
    \addtocounter{footnote}{-1}%
    \endgroup
}

\definecolor{myblue}{RGB}{20,80,150}  
\definecolor{myred}{RGB}{160,30,30}   
\definecolor{mygreen}{RGB}{50,120,50} 

\title{Structural priors for data-efficient language learning}

\author{
    Yana Veitsman\textsuperscript{*} ~~~~
    Jonas Mayer Martins\textsuperscript{*} ~~~~
    Jonathan Lautenschlager ~~~~
    Lisa Beinborn \\
    University of G\"{o}ttingen, Germany \\ 
    \texttt{firstname.lastname@uni-goettingen.de}}

\begin{document}

\maketitle
\blfootnote{$^*$~These authors contributed equally to this work.}

\begin{abstract}
Efficient language learning requires methods to reduce the reliance on large data and computational resources. We investigate \emph{structural transfer}: First training models on non-language data to induce useful priors for natural language. This approach is a form of weight initialization for multilingual language modeling. We evaluate transfer via next-token-prediction loss, weight shifts in the model, and downstream linguistic benchmarks. Several symbolic data types---notably music, probabilistic grammars, and cellular automata---yield lower language-modeling loss than random initialization. These gains coincide with smaller weight shifts during subsequent language training, suggesting that structural transfer positions models in a more favorable region of the parameter space. However, a lower loss does not translate consistently into better downstream linguistic performance, and transfer from non-language data is less efficient than additional language data. We conclude that non-language data can serve as a partial substitute for language data for the training objective of next-token prediction but does not reliably support broader linguistic generalization.
\end{abstract}

\begin{center}
\small
\href{https://hf.co/collections/huds-uni-goe/blm2026}{%
  \raisebox{-0.4em}{\includegraphics[height=1.3em]{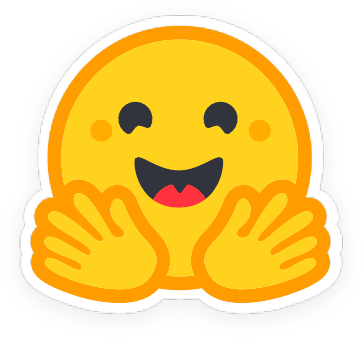}}
  Models and data} 
\hspace{8pt}|\hspace{7pt}
\href{https://gitlab.gwdg.de/huds/projects/blm2026}{%
  \raisebox{-0.33em}{\includegraphics[height=1.2em]{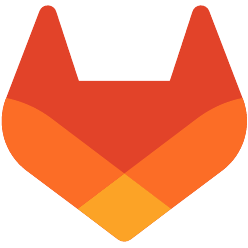}} 
  Code repository}
\end{center}

\section{Introduction}

The performance of language models scales strongly with data and compute, yet both resources are fundamentally limited. Both modeling low-resource languages and cognitively plausible modeling, for instance, require learning from little input, making data efficiency an engineering problem and a scientific challenge \cite{hoffmann-etal-2022-training}. The BabyLM shared task addresses this constraint directly by evaluating systems under low-data conditions and encouraging cognitively inspired approaches \cite{choshen-etal-2026-babylm}.

A natural source of inspiration for data-efficient learning is human language acquisition. Despite exposure to relatively limited linguistic input, infants detect and exploit statistical regularities in the language they perceive \cite{saffran-kirkham-2018-infant}. We examine whether it is possible to introduce structural priors into language models that facilitate more data-efficient language learning. We therefore aim to understand to what extent non-language data give rise to representations that facilitate subsequent language learning.

Such priors can be instilled in language models through \emph{pre-pretraining}: an additional stage in which models are trained on non-language data before natural-language pretraining \cite{papadimitriou-jurafsky-2020-learning, hu-etal-2025-circuits, lee-etal-2026-training}. Pre-pretraining constitutes a form of structural transfer, in which a model first learns to predict a structured symbolic signal and then reuses the resulting parameter state during language training. Previous work suggests that this approach can improve language-modeling convergence and downstream performance in domains such as coding, mathematics, and reasoning \cite{lee-etal-2026-training,shinnick-etal-2025-transformers}. However, the relationship between the parameter changes induced by pre-pretraining and performance on the target objective remains underexplored. 

Data efficiency is particularly important in multilingual settings, where models must generalize to typologically and distributionally diverse languages with varying amounts of data. At the same time, multilingual training introduces challenges absent from monolingual settings, including cross-lingual interference from data mixtures \cite{ye-etal-2025-data} and conflicting optimization signals \cite{wang-etal-2023-gradsim}. The multilingual setting thus offers the opportunity to test not only whether gains from structural transfer in monolingual or task-specific settings persist, but also whether structural transfer is language-specific.

\paragraph{Approach and contributions}
We investigate whether structural transfer can accelerate multilingual language learning. We first train GPT-2-style models on structural data---including probabilistic context-free grammars, cellular automata, piano music, and protein sequences---and then continue training on the multilingual BabyLM corpus \cite{jumelet-etal-2026-babybabellm}. We evaluate transfer along three axes: 1) efficiency on the natural-language next-token prediction objective, 2) model-internal weight shifts, and 3) downstream performance under the BabyLM evaluation framework \cite{choshen-etal-2026-babylm}. This setup allows us to test whether certain types of structural data yield a weight initialization that benefits multilingual modeling.

\begin{figure*}
    \centering
    \includegraphics[width=\linewidth]{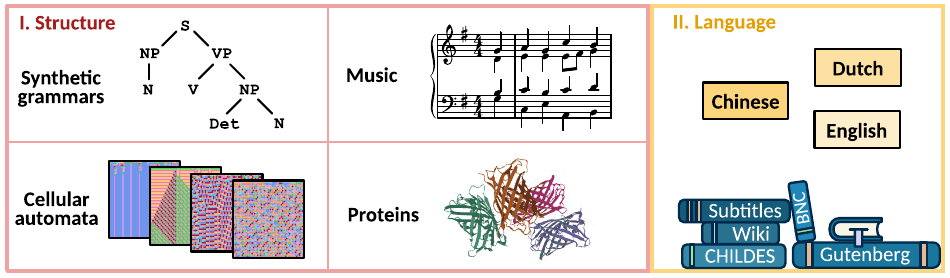}
    \caption{Schematic of the experimental setup. To investigate how training on structure transfers to learning language, we first train language models on one of four symbolic data types: 1) probabilistic context-free grammars; 2) cellular automata; 3) piano music; and 4) protein sequences. The example tree could refer to the sentence \emph{Alice paints the cat.} The piece shown is the beginning of \citet{reger-1900-mond}. The protein depicted is green fluorescent protein (GFP) of jellyfish adapted from the \href{https://www.uniprot.org/uniprotkb/P42212/feature-viewer}{Uniprot entry} \cite{boutet-etal-2007-uniprotkb}. Afterwards, we train these models on a multilingual corpus comprising Chinese, Dutch, and English texts.}
    \label{fig:graphical-abstract}
\end{figure*}


\section{Related work}
\label{sec:related-work}
Small-scale language models are motivated by both practical constraints, such as limited data and compute, and cognitively plausible learning scenarios. Work in this setting has shown that models can benefit from structural biases introduced during training \cite{papadimitriou-jurafsky-2020-learning,papadimitriou-jurafsky-2023-injectinga,hu-etal-2025-circuits,lee-etal-2026-training}. We view such biases through the lens of initialization: prior exposure to structural data can move a model away from an uninformed random starting point and toward a parameter region that supports more efficient language learning.

\subsection{Sample-efficient modeling}
Previous entries in the BabyLM challenge \cite{choshen-etal-2026-babylm} have explored a range of mechanisms for improving sample efficiency, including learning from interaction \cite{charpentier-etal-2025-babylm,mayermartins-etal-2025-once}, curriculum learning \cite{warstadt-etal-2023-findings,diehlmartinez-etal-2023-climb}, and multilingual learning contexts \cite{jumelet-etal-2026-babybabellm}. These approaches share the goal of improving learning when additional data or compute cannot be scaled up.

\paragraph{Multilingual modeling}
Studying how models generalize across diverse linguistic distributions and data scarcity is especially relevant in multilingual settings. An open question is whether performance disparities across languages reflect model or data biases, or instead arise from specific linguistic properties \cite{mielke-etal-2019-what,poelman-etal-2025-confounding,shani-etal-2026-roots}. Multilingual training requires careful design decisions about data mixtures \cite{ye-etal-2025-data} and optimization across languages \cite{wang-etal-2023-gradsim}. One line of work suggests that transfer is most effective between languages with similar linguistic properties \cite{pires-etal-2019-how,k-etal-2020-crosslingual,snaebjarnarson-etal-2023-transfer}; another suggests that similarity may matter less than matching the informational complexity of the training signal \cite{lee-etal-2026-training}. The BabyLM challenge offers a controlled setting for examining whether structural transfer can support learning of heterogeneous distributions.

\subsection{Weight initialization and inductive biases}
In neural language models, inductive biases can be introduced not only through architecture, but also through the model parameter state. Classical initialization schemes use random weights with carefully chosen scaling coefficients to break symmetry and stabilize activations and gradients \cite{rumelhart-etal-1986-learning,glorot-bengio-2010-understanding,he-etal-2015-delving}. Although these schemes do not encode task-specific structure, they shape the starting conditions for optimization. Structural pretraining extends this idea: rather than relying only on random initialization, a model can first be trained on a controlled signal that may induce representations useful for later language learning \cite{papadimitriou-jurafsky-2023-injectinga,hu-etal-2025-circuits,lee-etal-2026-training}.

\subsection{Structural transfer}
We understand structural transfer as an umbrella term, encompassing the closely related concepts of curriculum learning, transfer learning, meta-learning, and pre-pretraining. In meta-learning, prior training across related tasks supports rapid adaptation; in curriculum learning, examples are ordered so that simpler data prepares the model for harder cases; and in transfer learning, knowledge from source tasks or languages provides a useful starting point for zero-shot performance or target-task fine-tuning \cite{wang-etal-2021-survey,lee-etal-2022-meta,soviany-etal-2022-curriculum}. Pre-pretraining on structural data \cite{papadimitriou-jurafsky-2020-learning,hu-etal-2025-circuits,lee-etal-2026-training,jiang-etal-2026-procedural} combines aspects of these approaches. It introduces an earlier training stage, often on structural non-language data to induce a useful initialization for subsequent language training. The relevant question is therefore not only whether structural data can be learned, but whether the induced parameter state is beneficial for natural language training.

\paragraph{Empirical findings.}
Structural data can vary in entropy, distributional shape, compositionality, hierarchy, and other regularities, which may induce different biases. Previous studies yield mixed conclusions about their benefits. Training on abstract grammars such as PCFGs can give rise to functional hierarchical representations \cite{jumelet-zuidema-2023-transparency, rohweder-etal-2026-hierarchical}, but downstream tasks are not always comparable: \citet{hu-etal-2025-circuits} report limited gains on BLIMP, while \citet{lee-etal-2026-training} focus more on reasoning benchmarks. It remains an unanswered question which properties of the structural source interact with the target language data and how they influence the performance on the evaluation task. We therefore examine a range of structural data types with varying degrees of abstractness and local and global dependencies. In our setup, cellular automata provide local rule-based dynamics, synthetic grammars contribute hierarchical and compositional structure, music offers sequential structure with long-range dependencies, and protein biological complexity.


\section{Data}
\label{sec:data}

The experimental setup consists of training a language model in two stages.
In stage~I, we train the model from scratch on a structural data type. In stage~II, we switch to training on the target corpus.

\subsection{Structural data}

In stage~I, we explore four structural data types: synthetic grammars, cellular automata, music, and protein sequences. As a baseline, we use unstructured sequences of random numbers.

\paragraph{Synthetic grammars} A probabilistic context-free grammar (PCFG) is a formal grammar in which production rules for non-terminal symbols are insensitive to the context \cite{manning-schutze-2000-probabilistic}. Every production rule is applied with a certain probability, making these synthetic grammars suitable for randomly sampling a wide variety of synthetic data. Created PCFGs mimic natural language closely by synthesizing grammar rules from the Penn Treebank \cite{marcus-etal-1993-building}, see \cref{sec:app:synthetic-grammars} for details. We test two synthetic datasets generated by the same grammar, differing only in the lexicalization: For \textsc{PCFG\textsubscript{Zipf}}, tokens follow a Zipfian distribution; for \textsc{PCFG\textsubscript{uni}}, tokens are distributed uniformly.

\paragraph{Cellular automata (CA)}

A cellular automaton is a row of cells, with each cell being in a discrete state. This row evolves over time under a local \emph{transition rule}~$f$ defining a set of update patterns. At each step in time, every cell is updated based on itself and its two neighbors. Cellular automata are a canonical example of self-organized complexity emergent from simple, deterministic rules \cite{vonneumann-burks-1966-theory,gardner-1970-mathematical,wolfram-1983-statistical}. The training objective requires the model to infer these rules from diverse patterns, making cellular automata a compelling source for learning functional capabilities. \Cref{sec:app:ncas} provides a visualization and technical details.

To assess the impact of structural complexity, we generate two dataset variants: \textsc{CA\textsubscript{16}} (medium-complexity dynamics with $K = 16$ states) and \textsc{CA\textsubscript{256}} (noisy dynamics with $K = 256$ states), each generated from about $1{,}820$ rules with multiple initial conditions per rule. The language model receives these space-time trajectories as concatenated rows. In training, the language model must infer the transition rule based on several trajectories generated from different initial conditions to succeed in next-token prediction.

\paragraph{Music}
Music is a highly structured non-language signal produced by humans. We use piano music from the \textsc{Aria-MIDI} dataset \cite{bradshaw-colton-2025-ariamidi}. For training, the language model receives a transcription of note events, timing and pedal state as a flat sequence of integers, see \cref{sec:app:music}. Musical data is characterized by long-range temporal dependencies with repetition and variations and rich hierarchical structure in the form of melody, harmony, and rhythm without encoding the semantics of natural language \cite{berezovsky-2019-structure}.

\paragraph{Protein sequences}
Proteins lie at the heart of vital biological functions and their amino-acid sequences encode highly complex molecular structures. We choose Swiss-Prot as a list of proteins \cite{boutet-etal-2007-uniprotkb} and normalize the dataset to the 20 canonical amino acids, each a single token, by replacing any rare amino acids or unknowns with an additional symbol~$X$.

\paragraph{Random}
As a control condition, we generate sequences with uniformly distributed length between 1 and $2{,}048$ from an alphabet of uniformly independently sampled integers $k\in\{0,\dots,512\}$.

\subsection{Language data}
\label{sec:language-data}
We use the dataset provided for the multilingual track of the BabyLM challenge. The dataset comprises cognitively plausible multilingual data according to the criteria outlined in \citet{jumelet-etal-2026-babybabellm}. The exact composition of the dataset is listed in \cref{sec:app:BLM_data}. For stage~II training, we sample English, Dutch, and Mandarin Chinese in proportion of 1/3 for each language.

\paragraph{Baselines}
Our primary baseline model is trained directly in stage~II from random initialization, without any prior exposure to structural data. We additionally train a secondary baseline on English Wikipedia \cite{wikimediafoundation-2023-english} during stage~I, in order to assess whether structural transfer provides benefits beyond those obtained from additional natural-language data.

\section{Experimental setup}
\label{sec:exp-setup}

We run experiments on 80/20 data splits with a \mbox{GPT-2} architecture \cite{radford-etal-2019-language}. Early stopping (with patience of 3 epochs and a threshold of 0.01) is used to identify the optimal amount of structural and natural-language data, respectively. For hyperparameters and implementation details, refer to \cref{sec:app:training-regime}. We evaluate the gains from structural transfer using the validation loss achieved by the trilingual model on a held-out portion of the natural-language corpus, together with performance on a combination of zero-shot and fine-tuning tasks from the BabyLM framework \cite{choshen-etal-2026-babylm}, see \cref{sec:app:BLM_evals}.

\paragraph{Tokenization}
We train a trilingual byte-pair-encoding (BPE) tokenizer with a fixed vocabulary of $16{,}897$ tokens using the same approach to data sampling as outlined in \cref{sec:language-data}. The tokenizer also includes 512 non-overlapping tokens used exclusively in stage~I. At the start of stage~II, we reinitialize the embedding layer to prevent spurious lexical transfer from the structural datasets. 

\paragraph{Loss metrics}
We measure the performance of the model on the next-token prediction through the cross-entropy loss~$\mathcal{L}(t)$ over the training tokens~$t$.

The average loss change relative to the baseline is defined by the \emph{loss ratio}~$\rho$ as the ratio of the areas below the loss curves up to $N$ tokens of a condition~$\mathcal{L}_i$ and the primary baseline~$\mathcal{L}_0$ for training on natural language, 
\begin{equation}
    \rho_i = \frac{\int_{t_\text{cutoff}}^{N} \mathcal{L}_i(t)\, \mathrm{d}t}{\int_{t_\text{cutoff}}^{N} \mathcal{L}_0(t)\, \mathrm{d}t} - 1\,,
\end{equation}
with a minimal cutoff at $t_\text{cutoff} = 2\,\mathrm{M}$, to avoid inflating the metric due to large loss differences early in training. 

A complementary metric, as used by \citet{hu-etal-2025-circuits}, is the \emph{token efficiency}~$\tau$, i.e., the ratio of tokens necessary to achieve the final baseline loss with structural transfer,
\begin{equation}
    \tau_i = \frac{1}{N} \bigg(N_i + \argmin_{0\leq n \leq N} \{\mathcal{L}_i(n) \leq \mathcal{L}_0(N)\}\bigg),
\end{equation}
where $N_i$ is the number of steps we train on structural data in stage~I for condition~$i$. Intuitively, this captures how many tokens are necessary to train on structural plus language data to beat the baseline.


\section{Results}
\label{sec:results}

We examine the effect of structural transfer from each condition along three dimensions: 1) next-token prediction loss on a multilingual corpus, 2) model-internal weight shifts during training, and 3) linguistic benchmarks. Together, these analyses allow us to assess to which extent structural transfer can aid language-model training.

\subsection{Transfer on next-token prediction}

\begin{figure}[htbp]
    \centering
    \includegraphics[width=\linewidth]{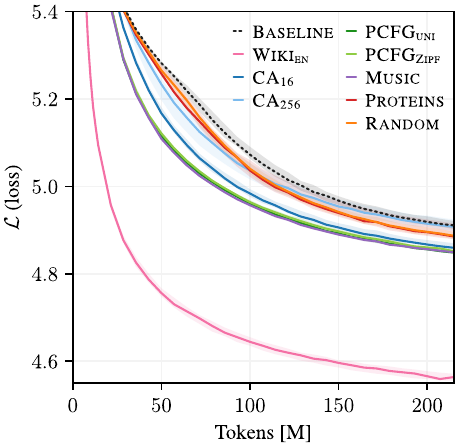}
    \caption{Cross-entropy loss $\mathcal{L}$ by training step of language models after having trained on structural data compared with the primary baseline of not training on any structure. The lines show the median over five seeds, except few runs that triggered early stopping. Shaded regions delineate minimum and maximum seed at each token count. Note that \textsc{PCFG\textsubscript{uni}}, \textsc{PCFG\textsubscript{Zipf}}, and \textsc{Music} overlap.}
    \label{fig:loss}
\end{figure}

Training on symbolic data can robustly improve the loss $\mathcal{L}$ on next-token prediction of natural language. \Cref{fig:loss} shows that every condition matches or outperforms the baseline of not training on any structure (dashed black) in terms of achieving a lower loss. The clearest gains come from \textsc{Music}, both synthetic grammars \textsc{PCFG\textsubscript{uni}} and \textsc{PCFG\textsubscript{Zipf}}, and the cellular automata \textsc{CA}\textsubscript{16}. Note that the two PCFG loss curves overlap, which implies that the Zipfian distribution does not benefit the structural transfer more than a uniform distribution. Although \textsc{CA}\textsubscript{256} yields an early advantage over the baseline, this condition converges to the baseline at 220\,M tokens, suggesting that the noisier dynamics do not induce an equally useful initialization. \textsc{Proteins} and \textsc{random} lead to more modest but consistent improvements.

Overall, these results indicate that next-token-prediction loss can be robustly improved with a variety of symbolic data. However, the \textsc{Wiki\textsubscript{en}} baseline puts these gains into perspective, showing that training on more natural language data for just one of the three languages from the multilingual corpus outperforms any of these symbolic data by a wide margin.

\subsection{Model-internal representations}

\begin{figure}[htbp]
    \centering
    \includegraphics[width=\linewidth]{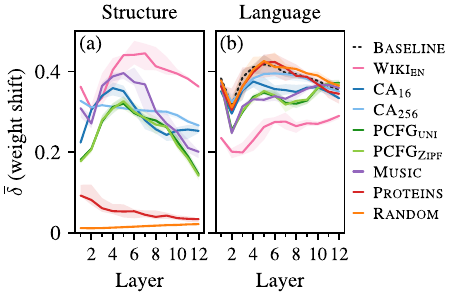}
    \caption{Weight shift~$\bar{\delta}$ per layer in the self-attention matrices when training on (a)~structure and subsequently (b)~language. Median seed with shaded regions for minimum and maximum seed.}
    \label{fig:weight_shift}
\end{figure}

To understand how structural transfer impacts the model, we examine how much the weights shift in the two stages of training on structure and language, respectively. The relative weight shift from a time~$t$ before to after the respective training stage for layer~$l$, component~$c$, and seed~$s$ is 
\begin{equation}
    \delta^{(t)}_{l,c,s} = \frac{\|W^{\text{after}}_{l,c,s} - W^{\text{before}}_{l,c,s}\|_\mathrm{F}}{\|W^{\text{before}}_{l,c,s}\|_\mathrm{F}}\,,
\end{equation}
where $\|\cdot\|_\mathrm{F}$ is the Frobenius norm. We plot the mean across components~$c$ of the self-attention mechanism in the transformer architecture (query, key, value, and output matrices) per layer,
\begin{equation}
    \bar{\delta}^{(t)}_{l,s} = \frac{1}{4} \sum_c \delta^{(t)}_{l,c,s}\,.
\end{equation}

\paragraph{Shifting weights}

In \cref{fig:weight_shift}~(a), showing stage~I, we observe that \textsc{proteins} and \textsc{random} induce much smaller weight shifts than the other better-performing conditions of stage~II, which indicates that a random initialization is close to a local optimum for predicting these signals.  Panel~(b) shows how much the weights shift during stage~II (language modeling). We see that training on sequences of \textsc{proteins} and \textsc{random} numbers in stage I does not reduce the required weight shifts for modeling language compared to the baseline of no structural training. The weights of models trained on synthetic grammars, \textsc{music}, and \textsc{CA}\textsubscript{16} change less than the baseline with an average weight shift of around $\bar{\delta} = 0.34$ across layers. 
Notably, \textsc{CA}\textsubscript{256} is an exception. The weights change strongly in the structural-learning phase, but the resulting attention pattern is not beneficial for language modeling, that is, the required weight shift is as large as for the randomly initialized baseline. 

\paragraph{Relating weights and loss}

\begin{figure}[tbp]
    \centering
    \includegraphics[width=\linewidth]{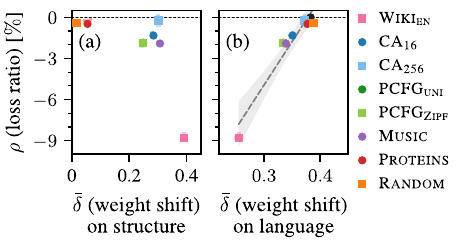}
    \caption{Relation of loss ratio on language data and weight shift, averaged over layers, for training on (a) structural data and (b) language data. Solid dots show the median, small transparent dots show each seed. \textsc{PCFG} and \textsc{music} perform best because the representations from learning on structure can be partially transferred to language, \textsc{CA}\textsubscript{16} to a lesser extent, too. Dotted black line and black circle (b): baseline of training on language only. Dashed gray line in (b): ordinary-least-squares regression through condition-level medians; shaded band: $95\,\%$ confidence interval.}
    \label{fig:shift-vs-efficiency}
\end{figure}

In \cref{fig:shift-vs-efficiency}, we visualize the loss reduction relative to the weight shift. The respective weight shift correlates with loss improvement on language (panel~(b) with Pearson correlation $r = 0.96$, $p < 0.01$) but less so on structural data (panel~(a) with $r = -0.60$, $p = 0.11$). The smallest weight shift achieved in stage~II is associated with the biggest loss improvement for the English Wikipedia baseline. These results indicate that the weights optimized for predicting structural data facilitate learning language by requiring a smaller weight shift. 

Moreover, the best conditions achieve a token efficiency of approximately $60\,\%$ of the original token budget, meaning they require only about $60\,\%$ of the baseline token budget to reach equivalent language-modeling performance. The loss ratio $\rho$ measures how much the training on structural data improves the cross-entropy loss (lower is better) on average over training steps and this ratio is closely related to token efficiency, see \cref{sec:app:metric_correlation}.

\subsection{Linguistic benchmarks}

\begin{figure*}[htbp]
    \centering
    \includegraphics[width=\linewidth]{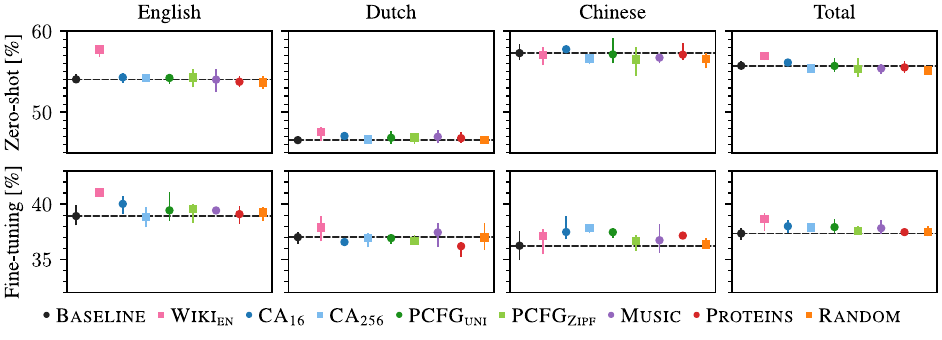}
    \caption{Average accuracy on linguistic zero-shot and fine-tuning benchmarks for English, Dutch, Chinese, and averaged over these three languages, see \cref{tab:app:evaluation_setup}. The dashed line denotes the baseline of not training on structural data. There are no significant changes relative to the baseline for the total, except \textsc{Wiki\textsubscript{en}} and \textsc{CA\textsubscript{16}}. Dots show the median and error bars indicate the minimum and maximum seed.}
    \label{fig:blm_eval}
\end{figure*}

We employ the framework of the BabyLM challenge, which combines zero-shot and fine-tuning evaluation in English, Dutch, and Chinese \cite{choshen-etal-2026-babylm}, covering testing suites for linguistic competence (BLiMP-style tasks), reading comprehension, knowledge probing, and others. For details on the exact tasks, refer to \cref{tab:app:evaluation_setup} in \cref{sec:app:BLM_evals}.

The results show that the observed loss reduction on the language modeling objective does not translate to better performance on linguistic benchmarks. \Cref{fig:blm_eval} shows the evaluation results for several zero-shot and fine-tuning tasks. The only stage-I condition that robustly improves the baseline results for English and Dutch is training on English texts. All other structural-transfer conditions do not significantly affect the average evaluation results positively nor negatively. 

Across languages, \textsc{CA\textsubscript{16}} has a small positive effect. We note that some conditions improve or worsen the performance exclusively on fine-tuning or zero-shot. This inconsistency might indicate that the linguistic benchmarks cannot resolve the small differences induced by the structural data or that zero-shot and fine-tuning tasks are not entirely aligned, measuring different linguistic aspects.
Even if some structural data tends to lead to an improvement, the magnitude of this effect appears small, on the order of one percentage point. 
We conclude that an improvement on the target objective (next-token prediction) does not generally entail an improvement on linguistic benchmarks. 

\paragraph{Detailed effects}

We fit a Bayesian hierarchical model to estimate the effect of each structural-transfer condition relative to the natural language baseline. Benchmark scores (averaged across seeds) are modeled as $\text{score}_{ctl} \sim \mathcal{N}(\mu_c + u_t, \sigma)$ for each language and as $\text{score}_{ctl} \sim \mathcal{N}(\mu_c + u_t + u_l, \sigma)$ for the total, with a fixed effect $\mu_c$ per condition, a random effect $u_t$ per task, and a random effect per language $u_l$. We report the posterior mean difference $\Delta_c = \mu_c - \mu_\text{baseline}$ relative to the effect of the primary baseline in percentage points with $95\,\%$ highest-density intervals (HDI) and posterior probability $p(\Delta_c > 0)$, estimated via NUTS sampling \citep{hoffman-gelman-2014-nouturn,abril-pla-etal-2023-pymc} with four chains of 2{,}000 draws.

\begin{figure}[htbp]
    \centering
    \includegraphics[width=\linewidth]{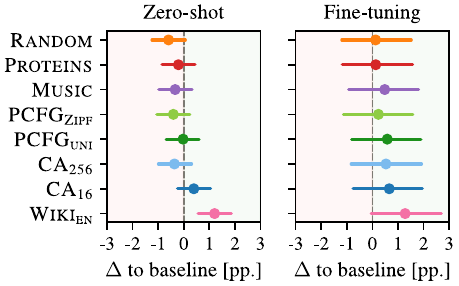}
    \caption{Mean and $95\,\%$ highest-density intervals estimated by a Bayes model for linguistic benchmarks of zero-shot and fine-tuning across languages.}
    \label{fig:bayes_model}
\end{figure}

\Cref{fig:bayes_model} shows posterior estimates for all conditions across languages combined. Training on Wikipedia in stage~I has robust positive effects on both zero-shot ($\Delta = +1.2$\,pp, HDI: $[0.6, 1.9]$, $p = 1.00$) and fine-tuning ($\Delta = +1.3$\,pp, HDI: $[-0.1, 2.7]$, $p = 0.97$) evaluations. The \textsc{Random} condition shows a small negative effect on zero-shot performance ($\Delta = -0.6$\,pp, HDI: $[-1.2, 0.1]$, $p = 0.04$). The \textsc{CA\textsubscript{16}} has a small positive effect on both zero-shot and fine-tuning performance ($\Delta = 0.4$\,pp, HDI: $[-0.2, 1.1]$, $p = 0.88$ and $\Delta = 0.7$\,pp, HDI: $[-0.8, 2.0]$, $p = 0.83$, respectively). All structural data have effects close to zero with HDIs that include zero.

Based on these results, we performed a hyperparameter search to train the submission model for the BabyLM challenge on a mixture of the most beneficial data types, see \cref{sec:app:challenge-submission} for details. However, even the best model (in terms of validation loss) does not perform much better than the separately trained models.

\paragraph{Syntactic Tasks} We hypothesize that structural transfer is more beneficial for syntax-level tasks due to its hierarchical dependencies. To check this, we select tasks explicitly marked as \enquote{syntactic} from English BLiMP \cite{warstadt-etal-2020-blimp}, BLiMP-NL \cite{suijkerbuijk-etal-2025-blimpnl} and ZhoBLiMP \cite{liu-etal-2026-systematic}. These tasks span syntactic features such as argument structure, ordering in wh-clauses, islands constraints, etc. However, both per-task and aggregated category results per data type do not improve over the baseline by more than two percentage points.


\section{Discussion}
\label{sec:discussion}

In summary, our results indicate that learning structural patterns enables faster performance gains on the next-token prediction task. However, these dynamics do not translate to better performance in linguistic benchmarks~\cite{chen-etal-2025-scaling}. Rather, transfer depends on the task at hand: Both next-token prediction and linguistic benchmarks measure language competence yet via different metrics, so that optimizing for next-token prediction does not necessarily imply measurable transfer effects on linguistic tasks. Finding that none of the tested structural conditions match the efficiency of additional natural-language data, we conclude that the intricate correlational structures of language cannot easily be replaced by synthetic data.

Given that the two PCFG variants perform equally, a Zipfian distribution alone does not appear to impact transfer significantly. This may be because a Zipfian distribution is a first-order approximation of the correlational structure, i.e., just unigrams, which are learned quickly when training on natural language. 

Although protein sequences encode highly complex biological information, this structure does not appear to be readily accessible to a transformer, at least not under our naive tokenization approach used for the next-token prediction task of one token per amino acid. Since even random numbers perform slightly better than a randomly initialized model without any structural training, this might indicate that even a very simple tweak to the weight initialization can improve training results.

The structural data may initialize the weights to functionally useful representations leading to successful structural transfer. These representations might be interesting to explore in connection with the work on discovering capabilities emerging in attention heads during pretraining \cite{olsson-etal-2022-incontext, elhage2021mathematical, crosbie-shutova-2025-induction, aoyama-etal-2026-predicting}, in which a learned matrix structure facilitates functions such as token copying or retrieval.

In some respects, the well-performing data are language-like. For example, music has highly complex structure produced by humans, including recursions and a form of syntax. PCFGs explicitly model syntactic structure extracted from natural-language texts. In cellular automata, on the other hand, complexity emerges from simple rules. However, any structural data we train on lacks the semantic content of language. Thus, it would be interesting to further explore the degree to which semantics are helpful and necessary for transfer. 


\section{Conclusion}
\label{sec:conclusion}
We investigated whether structural transfer can improve sample-efficient multilingual language modeling. Our results show that training on structured symbolic data before natural language can improve next-token prediction learning efficiency. The weight-shift analysis supports this interpretation: Models that benefit most from structural transfer require smaller parameter changes during multilingual language training, indicating that prior structural exposure places models in a more favorable region of the parameter space.

However, downstream gains are limited. Improvements on the language-modeling objective do not reliably translate into stronger performance on downstream linguistic benchmarks. Moreover, when additional natural language data is available, it remains more efficient than structural data, even when training on a mix of partially typologically distant languages such as English, Dutch, and Chinese. Although this may be useful when language data is limited, such as for low-resource languages or other data-constrained settings, structural transfer is not a direct substitute for relevant language input.

More broadly, our findings suggest that structural data provides a direction for revisiting initialization strategies in language models. Carefully tailoring model weights before training may provide a useful head start, particularly in the context of limited compute and model capacity.

\section*{Limitations}

\paragraph{Selected structures} Although we aim for a wide range of structural data (synthetic grammars, music, proteins, CAs) that we expected a priori to possibly yield strong transfer to natural language, resembling language to varying degrees, we cannot account for all possible data types exhaustively.

\paragraph{Model size} We conduct our experiments on the GPT-2-small architecture. While this setup provides us with a reasonable overview of structural transfer, a variation of the model size could clarify the limits of the loss improvements on next-token prediction.

\paragraph{Data-mixing effects} We do not exhaustively explore the effects of data mixing during stage~I. A wider variety of high-quality data in stage~II would allow for a more detailed study of language-specific differences in structural transfer. Since the Wikipedia baseline is in English only, we cannot assess the benefits of structural transfer across different natural languages. Observing effects of structural transfer tailored to a specific language would be an interesting avenue for future research.

\paragraph{Representations} We evaluate transfer by analyzing the magnitude of the weight shift. A more detailed analysis through the lens of mechanistic interpretability could target specifically the representations that the model learns on the structural data and possibly reuses on natural language.

\section*{Ethical considerations}

While we aim to advance the design of cognitively inspired models, we do not speculate here on how human language processing works. Explicit parallels between the processing capabilities of models and human cognition tend to be strained easily.

\section*{Acknowledgments}

We thank the reviewers for their feedback. We thank Alexander Ecker and Edoardo Ponti for helpful discussions. This research is partially supported by the zukunft.niedersachsen program of the VolkswagenStiftung (L.B., Y.V.) and by a VENI grant (Vl.Veni.211C.039) from the Nederlandse Organisatie voor Wetenschappelijk Onderzoek (NWO) (L.B.).

\section*{Author contributions}

\textbf{Y.V.:} Supervision, Conceptualization, Analysis, Software (training), Literature review, Writing---original draft (Abstract, Introduction, Related work, Conclusion, parts of Methods), Writing---review \& editing.
\textbf{J.M.M.:} Supervision, Conceptualization, Analysis, Software (data and analysis), Literature review, Visualization, Writing---original draft (Results and Discussion; parts of Introduction, Related work, and Methods; App.\ on CAs and music), Writing---review \& editing.
\textbf{J.L..:} Software (PCFG data), Writing---original draft (PCFGs).
\textbf{L.B.:} Supervision, Conceptualization, Analysis, Writing---review \& editing.

\bibliography{refs}


\appendix

\section{Synthetic grammars}
\label{sec:app:synthetic-grammars}

\begin{figure}[htbp]
    \centering
    \includegraphics[width=\linewidth]{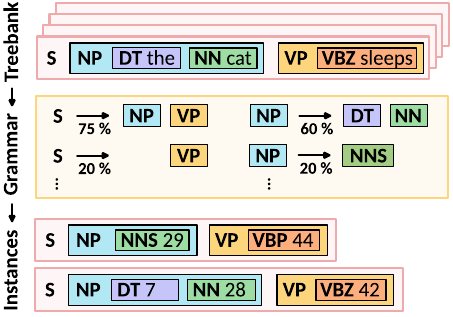}
    \caption{From the treebank, which consists of sentences with part-of-speech tags, we derive the transition probabilities, e.g., a sentence \texttt{S} splits into a noun phrase \texttt{NP} and a verb phrase \texttt{VP} or into only a verb phrase. Similarly, a noun phrase can split into a determiner \texttt{DT} and a noun \texttt{NN} or into a plural noun \texttt{NNS}. From the derived grammar, we can sample instances again, but the lexicalization is not semantic, represented by abstract integers instead. The model receives only the sequence integers, without part-of-speech tags.}
    \label{fig:app:pcfg_schematic}
\end{figure}

We induce a probabilistic context-free grammar from the UD-Treebank annotation. The grammar is delexicalized, i.e., the terminal symbols are word-class categories. We then create data by sampling parse trees from the grammar and instantiating word classes with vocabulary IDs, see \cref{fig:app:pcfg_schematic}. The two types \textsc{PCFG\textsubscript{uni}} and \textsc{PCFG\textsubscript{Zipf}} differ only in the frequency with which the words are sampled in the last step. \Cref{tab:app:pcfg_parameters} summarizes shared settings; \cref{tab:app:lexicon_layout} details the vocabulary layout.

%

\begin{table}[htbp]
    \small
    \centering
    \begin{tabular}{ll}
    \toprule
    \textbf{Parameter} & \textbf{Value} \\
    \midrule
    Corpus size & 200\,M tokens \\
    Grammar & Penn Treebank PCFG \\
    Max. parse depth & 16 \\
    Sampling (PCFG\textsubscript{zipf}) & Zipf ($\alpha=1$) \\
    Sampling (PCFG\textsubscript{uni}) & uniform \\
    \bottomrule
    \end{tabular}
    \caption{PCFG hyperparameters.}
    \label{tab:app:pcfg_parameters}
\end{table}
\paragraph{English-based Grammar Induction and Normalization.} We derive a PCFG from the Wall-Street-Journal portion of the English Penn Treebank that is distributed with NLTK \cite{marcus-etal-1993-building,bird-loper-2004-nltk} by estimating relative frequencies. The grammar is entirely delexicalized during construction by removing word-level rules, resulting in 2{,}664 structural productions, 26 non-terminal symbols, and 22 terminal categories. Part-of-speech tags on the right-hand side are collapsed into coarse abstract categories, such as grouping various noun tags into a single noun category. Punctuation is mapped to dedicated terminal categories (e.g. \texttt{period}, \texttt{comma}), rather than being collapsed into a single punctuation class. Functional tags and trace indices are stripped from non-terminal names to merge similar Treebank labels, and productions containing empty nodes are excluded.

After merging duplicate productions, the probabilities are normalized separately for each left-hand-side non-terminal. To guarantee that generation can terminate at a specified depth, some non-terminals require a fallback rule.
For any non-terminal lacking a production that expands exclusively to terminal categories, we add a single fallback production with a raw probability of $10^{-6}$. This fallback is included in the left-hand side group prior to normalization, allowing it to retain a small relative probability while slightly reducing the mass of the other rules.

\paragraph{Tree generation and depth constraints.}
Parse trees are generated via stochastic top-down sampling from the induced PCFG, starting at the root symbol \texttt{S}. 
At each non-terminal node, a production is drawn with a probability proportional to its weight among the rules for that symbol, and expansion continues until every leaf is a terminal category. 
For example,\\[0.25em]
\begin{tabular}{@{}r@{\;}c@{\;}l@{}}
\texttt{S}  & $\rightarrow$ & \texttt{NP}\ \texttt{VP} \\
\texttt{NP} & $\rightarrow$ & \texttt{det}\ \texttt{noun} \\
\texttt{VP} & $\rightarrow$ & \texttt{verb}\ \texttt{NP} \\
\texttt{NP} & $\rightarrow$ & \texttt{noun} \\[0.2em]
            & $\Rightarrow$ & \texttt{det}\ \texttt{noun}\ \texttt{verb}\ \texttt{noun}
\end{tabular}\\[0.35em]
To keep the trees finite, we enforce a maximum depth limit of 16, counting the root as depth 0. During expansion at depths 0 through 14, any available production for a given symbol may be used. When the sampler reaches depth 15, the active rule set is restricted strictly to terminal-only productions. This intervention truncates very deep structures and slightly shifts the generated distribution compared to the unbound grammar.

\paragraph{Lexicon layout and token sampling}
The 512 integer IDs are shuffled using a fixed random seed and partitioned into disjoint slices for each syntactic category, see \cref{tab:app:lexicon_layout}. Slot sizes are derived from Penn Treebank \emph{type} counts (distinct word forms per category), not from token frequencies in running text. 
To account for the fact that only a small portion of the vocabulary serves as function words, we unevenly distribute the 512 vocabulary IDs into 74 slots ($\approx 15\,\%$) for closed-class terminals and 438 slots for open-class terminals.

Once the tree is fully expanded, each terminal leaf is mapped to a concrete ID from the respective vocabulary slice. Under a Zipfian condition, tokens within a slice are sampled with weights proportional to their inverse rank using a global exponent of $\alpha = 1$. Under the uniform condition, every token within the slice has an equal selection probability.

\begin{table}[htbp]
    \small
    \centering
    \begin{tabular}{ll}
    \toprule
    \textbf{Parameter} & \textbf{Value} \\
    \midrule
    Number of categories & 22 (disjoint slices) \\
    Vocabulary size & 512 opaque IDs \\
    Open / closed budget & 438 / 74 (85\,\%\,/\,15\,\%) \\
    Allocation within budgets & PTB type counts \\
    Largest closed slice & num (51) \\
    Within-slice sampling & Zipf ($\alpha{=}1$) or uniform \\
    \bottomrule
    \end{tabular}
    \caption{Lexicon layout.}
    \label{tab:app:lexicon_layout}
\end{table}

\paragraph{Final Output and Corpus Statistics.} Each generated tree yields one sequence of tokens by extracting the terminal IDs in left-to-right order. The final output consists of simple integer sequences ranging from 0 to 511. We stop the generation process at exactly $200{,}000{,}000$ tokens per corpus, possibly truncating the last sentence. This procedure results in $572{,}648$ sentences for each file and a mean sequence length of about 349.

\section{Generating cellular automata}
\label{sec:app:ncas}

\begin{figure}[htbp]
    \centering
    \includegraphics[width=\linewidth]{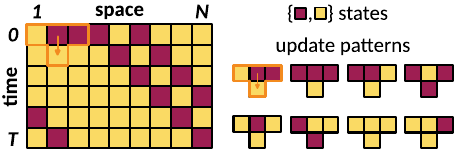}
    \caption{A cellular automaton consists of a row of cells, each in one of, e.g., $K=2$ states $\{\textcolor[HTML]{9e1e52}{\blacksquare}, \textcolor[HTML]{f9d95b}{\blacksquare}\}$. The next time step is determined by applying a set of update patterns that define the \emph{transition rule} of the cellular automaton to that row.}
    \label{fig:app:diagram_nca_trajectory}
\end{figure}

To illustrate the mechanics of cellular automata, consider the transition rule in \cref{fig:app:diagram_nca_trajectory}: The update pattern $f(\textcolor[HTML]{f9d95b}{\blacksquare}\, \textcolor[HTML]{9e1e52}{\blacksquare}\, \textcolor[HTML]{9e1e52}{\blacksquare}) =  \square \textcolor[HTML]{f9d95b}{\blacksquare} \square$ maps this three-cell neighborhood to the state $\textcolor[HTML]{f9d95b}{\blacksquare}$ in the next time step. 

The rule space of classical cellular automata with $K$ states consists of ${K^K}^3$ rules, which is astronomically large even for $K = 16$, and many of the rules produce either fixed points or uncorrelated noise. Rather than sampling this space uniformly, we parameterize transition rules through a small neural network $f_\theta$. Each set of randomly sampled, frozen parameters~$\theta$ defines a rule. 

This approach is known as neural cellular automata (NCA) \cite{mordvintsev-etal-2020-growing, spitznagel-keuper-2026-new}. By defining a suitable probability distribution for sampling the network parameters $\theta$, we avoid both entirely noisy and homogeneous dynamics, that are expected to be less beneficial for structure learning. Our setup of using NCA for structural transfer in language models is inspired by \citet{lee-etal-2026-training} but uses a one-dimensional grid, a simplified architecture, and deterministic (argmax) updates for efficiency and interpretability.

\begin{figure}[htbp]
    \centering
    \includegraphics[width=\linewidth]{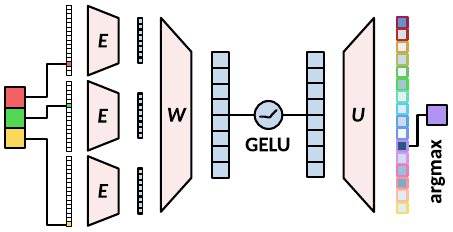}
    \caption{Schematic of the NCA architecture $f_\theta$, which takes a three-cell neighborhood as input and outputs the next state for the middle cell. The middle cell in state $\textcolor[HTML]{52d752}{\blacksquare}$ is updated based on itself and its neighbors $\textcolor[HTML]{f26262}{\blacksquare}$ and $\textcolor[HTML]{f9d95b}{\blacksquare}$ through the following steps: One-hot encoding of the state, projection $E$ of this vector to a low-dimensional representation, then a convolution with a matrix $W$, applying a nonlinearity (GELU), unembedding $E$ and taking the argmax over this vector to arrive at the next cell state $\textcolor[HTML]{52d752}{\blacksquare} \to \textcolor[HTML]{b093f2}{\blacksquare}$.}
    \label{fig:app:diagram_nca_architecture}
\end{figure}

Every NCA still defines a classical CA rule. Given the fixed parameters $\theta$, one can simply enumerate the output of the NCA $f_\theta$ for all the three-cell neighborhoods to generate a classical lookup table. The neural parameterization is thus purely used to steer the data generation, not changing the nature of the simulated CAs but only which rules are likely to be drawn.

\subsection{Technical implementation}
In our implementation, each rule is defined by a randomly sampled, fixed parameter set~$\theta$, see \cref{tab:app:nca_initialization}. These parameters define a function $f_\theta$ that is applied to every cell and its $3$-cell neighborhood with periodic boundary conditions. This function is a neural network, which updates a cell $i$ to the state $\smash{c_i^{(t)}}$ at time $t$.

\paragraph{Architecture} The architecture for generating NCA is a simple sequence of three functional steps, which suffice to give rise to interesting dynamics by inducing a bottleneck, convolution, and nonlinearity on the input data. \Cref{fig:app:diagram_nca_architecture}  visualizes this architecture schematically. 

First, the cell states (a one-hot encoding of the $K$ integers, i.e., the vocabulary) are projected into a low-dimensional representation with an embedding matrix $E \in \mathbb{R}^{d \times K}$, such that
\begin{equation}
    \vec{e}_j = E \vec{1}_{c_j^{(t)}}
\end{equation}
for $j \in \{i-1, i, i+1\}$.

Second, these three embedding vectors $\vec{e}_j$ are concatenated. A matrix $\vec{W}$ and bias $\vec{b}$ act as a convolution of the three cell states, followed by a GELU nonlinearity to arrive at a vector
\begin{equation}
    \vec{h}_i = \mathrm{GELU}(W [\vec{e}_{i-1}\|\vec{e}_{i} \|\vec{e}_{i+1}] + \vec{b})
\end{equation}
where $\|$ means concatenation of the vectors. Finally, we apply an unembedding matrix $U$ to arrive at a distribution over $K$ states and take the argmax to select an output state,
\begin{equation}
    c_i^{t+1} = \argmax_{k\in K}(U \vec{h}_i)\,.
\end{equation}
The rules are thus fully deterministic given the parameters and an initial condition. For an initial condition $\vec{c}^{(0)}$, the update rule $\vec{c}^{(t+1)} = f_\theta(\vec{c}^{(t)})$ defines a trajectory $\tau = \big(\vec{c}^{(0)}, \vec{c}^{(1)}, \dotsc, \vec{c}^{(T-1)}\big)$ with $T$ time steps. The data that the language model receives is a row-wise flattened version of the trajectories $\tau$, separated by a token for the end of a sequence.

\paragraph{Generation}

For each dataset, we sample $100{,}000$ simulated trajectories from about $1{,}820$ rules. Each rule instantiates a neural cellular automaton. The parameters $\theta$ are sampled from distributions scaled by a parameter $s$ to elicit a variety of dynamical regimes,  see \cref{tab:app:nca_initialization}.
We generate two dataset variants, \textsc{CA\textsubscript{16}} and \textsc{CA\textsubscript{256}}, differing in the number of states $K$, embedding dimension $d$, and hidden dimension $h$, see \cref{tab:app:nca_architecture}. We tuned the hyperparameters for generating \textsc{CA\textsubscript{16}} manually to instill an initial bias favoring a medium range of entropy, which we hypothesize to be the most promising, cf.\ the spectrum of NCA visualized with four examples in \cref{fig:graphical-abstract}. The variant \textsc{CA\textsubscript{256}} contrasts this with a larger vocabulary and more noisy dynamics.

We also randomly vary the grid size~$N$, trajectory length~$T$, and number of trajectories per rule for more variety in the dataset, to prevent the model from overfitting on a specific periodicity in the data, see \cref{tab:app:nca_dataset_generation}. Each trajectory is initialized in a state $\vec{c}^{(0)} \in \{0, \dotsc, K-1\}^N$ by drawing uniformly from four schemes: (i) \emph{uni}, which draws each cell i.i.d.\ from the state set, $\mathcal{U}\{0, \dotsc, K-1\}$; (ii) \emph{gradient}, which interpolates linearly between two independently chosen states; (iii) \emph{block}, which creates regions of length $\ell \sim \mathcal{U}\{1, \lfloor N/7 \rfloor\}$, each with a single random state; and (iv) \emph{sparse}, which picks a uniform background state with a random fraction $\rho \sim \mathcal{U}(0.01,\,0.2)$ of cells replaced by random states.

\begin{table}[htb]
    \small
    \centering
    \begin{tabular}{llll}
    \toprule
     & \textbf{Parameter} & \textbf{Shape} & \textbf{Distribution} \\
    \midrule
    $s$                  & Weight scale       & scalar        & $\log(s)$ \\
                         &                    &               &  $\sim \mathcal{U}(\log(0.1),\; \log(3))$ \\
    $E$                  & Embedding          & $d \times K$  & $\mathcal{N}(0,\, s^2)$ \\
    $W$                & Conv.\ weights       & $h \times 3d$ & $\mathcal{N}(0,\, s^2 / (3d))$ \\
    $\mathbf{b}$       & Conv.\ bias          & $h$           & $\mathcal{N}(0,\, (0.2s)^2)$ \\
    $U$                & Unembedding        & $K \times h$  & $\mathcal{N}(0,\, s^2 / h)$ \\
    \bottomrule
    \end{tabular}
    \caption{Parameter initialization for each randomly sampled NCA rule~$\theta$. The weight scale~$s$ is shared across all parameters of a given rule and controls the dynamical regime of the resulting automaton.}
    \label{tab:app:nca_initialization}
\end{table}
\begin{table}[htb]
    \small
    \centering
    \begin{tabular}{llcc}
    \toprule
    & \textbf{Hyperparameter} & CA\textsubscript{16} & CA\textsubscript{256} \\
    \midrule
    $K$   & Number of states      & 16  & 256 \\
    $d$   & Embedding dimension   & 8   & 16  \\
    $h$   & Hidden dimension      & 8   & 6   \\
          & Neighborhood size     & 3   & 3   \\
    \bottomrule
    \end{tabular}
    \caption{Architecture hyperparameters for the two variants of the NCA dataset.}
    \label{tab:app:nca_architecture}
\end{table}
\begin{table}[htb]
    \small
    \centering
    \begin{tabular}{llc}
    \toprule
     & \textbf{Statistic} & \textbf{Value} \\
    \midrule
          & Number of rules  (expected)    & ${\approx}1{,}820$ \\
    $N$   & Grid size                      & $\mathcal{U}\{10, 100\}$ \\
    $T$   & Trajectory length              & $\mathcal{U}\{10, 100\}$ \\
          & Trajectories per rule          & $\mathcal{U}\{10, 100\}$ \\
          & Total trajectories             & $100{,}000$ \\
    \bottomrule
    \end{tabular}
    \caption{Statistics for generating the NCA datasets. Grid size~$N$, trajectory length~$T$, and trajectories per rule are sampled independently per rule.}
    \label{tab:app:nca_dataset_generation}
\end{table}

\section{Music}
\label{sec:app:music}

\begin{figure*}[htbp]
    \centering
    \includegraphics[width=\linewidth]{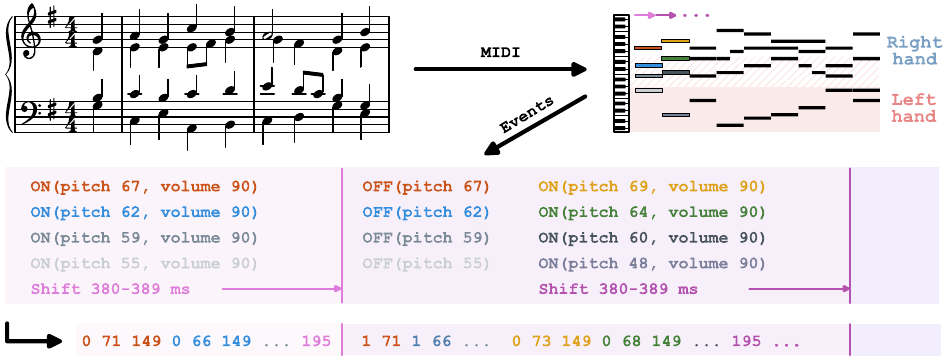}
    \caption{Piano pieces are encoded as MIDI format in the \textsc{Aria-MIDI} dataset. We extract the pitch, volume, pedal state, and shifts of the tape head and map each event to integers. This examples shows the beginning of \citet{reger-1900-mond}.}
    \label{fig:app:music_schematic}
\end{figure*}

We use the deduplicated \enquote{Unique} subset of the \textsc{Aria-MIDI} dataset \cite{bradshaw-colton-2025-ariamidi}, comprising $32{,}522$ piano pieces in MIDI format. We convert each piece into an integer sequence for the language model to train on.

\paragraph{Event encoding}

From the MIDI data, we extract three event types: 1)~\textsc{on} means a note begins (followed by two integers expressing pitch and velocity); 2)~\textsc{off} means a note ends (followed by an integer for the pitch, respectively); (3)~\textsc{pedal\_on} and \textsc{pedal\_off} mean the sustain pedal changes state. A \textsc{shift} token precedes events not simultaneous with the previous one, encoding elapsed time measured in bins of 10 milliseconds, effectively moving the tape head forward in time. Left and right hand are a single event stream.

\paragraph{Integer mapping}

Each event type above is assigned a contiguous integer range for transcribing the events into a sequence of numbers. The four command tokens \textsc{on}, \textsc{off}, \textsc{pedal\_on}, and \textsc{pedal\_off} are encoded as 0 to 3, respectively. Pitch (MIDI range 0--127) maps to the range 4--131. Velocity (MIDI range 0--127) is quantized in 25 steps of size 5, mapping to 132--156. Time shifts are binned in steps of 10 milliseconds into 127 bins, mapping to 157--283.
The full vocabulary thus contains $284$ tokens. \Cref{fig:app:music_schematic} illustrates the encoding on a short excerpt.

\section{BabyLM natural-language data}
\label{sec:app:BLM_data}

The multilingual training data from the BabyLM corpus is listed in \cref{tab:app:BLM_corpus}.

\begin{table}[htbp]
    \centering
    \small
    \begin{tabular}{ll S[table-format=2.2,table-number-alignment=right]}
        \toprule
        \textbf{Language} & \textbf{Domain} & \textbf{Ratio (\%)} \\
        \midrule
        \textbf{Chinese} & Conversational & 79\,\% \\
                         & Educational    & 9\,\%  \\
                         & Written/Books  & 12\,\% \\
        \midrule
        \textbf{Dutch}   & Conversational & 3\,\%  \\
                         & Educational    & 17\,\% \\
                         & Written/Books  & 16\,\% \\
                         & Subtitles      & 1\,\%  \\
                         & Fallback/Other & 63\,\% \\
        \midrule
        \textbf{English} & Conversational & 37\,\% \\
                         & Written/Books  & 42\,\% \\
                         & Fallback/Other & 21\,\% \\
        \bottomrule
    \end{tabular}
    \caption{Within-language domain mix of the BabyBabelLM Chinese, Dutch, and English corpora \cite{jumelet-etal-2026-babybabellm}. Total sizes: $138\,\mathrm{M}$, $110\,\mathrm{M}$, $99\,\mathrm{M}$ tokens, respectively. Percentages sum to 100\,\% per language.}
    \label{tab:app:BLM_corpus}
\end{table}

\section{Training regime}
\label{sec:app:training-regime}
We keep hyperparameters fixed across different structural data types and training stages, listed in the \cref{tab:app:training-hyperparameters}. Our Python environment uses \texttt{torch==2.12.0}, \texttt{transformers==4.47.0}, \texttt{datasets==3.0.0}, \texttt{flash\_attn==2.83}. All experiments were performed on a single NVIDIA A100 80GB GPU for a total of $\approx 250$ GPU hours.

\addtolength{\belowcaptionskip}{-10pt}
\begin{table}[htbp]
    \centering
    \small
    \begin{tabular}{ll}
    \toprule
    \textbf{Hyperparameter} & \textbf{Value} \\
    
    \midrule
    \multicolumn{2}{l}{\textit{Model}} \\
    \midrule
    Layers & 12 \\
    Hidden size & 768 \\
    Attention heads & 12 \\
    Context length & 512 \\
    Vocabulary size & 16{,}897 \\
    Activation & GELU \\
    Dropout & 0.1 \\
    LayerNorm $\epsilon$ & $1\times10^{-5}$ \\
    Initializer range & 0.02 \\
    
    \midrule
    \multicolumn{2}{l}{\textit{Training}} \\
    \midrule
    Sequence length & 512 \\
    Batch size & 16 \\
    Optimizer & AdamW \\
    Optimizer learning rate & $1\times10^{-4}$ \\
    Weight decay & 0.05 \\
    Warmup ratio & 0.01 \\
    Gradient clipping & 1.0 \\
    Early stopping patience & 3 \\
    Early stopping delta & 0.01 \\
    Seed & 11; 17; 42; 2{,}000; 3{,}407 \\
    
    \midrule
    \multicolumn{2}{l}{\textit{Fine-tuning}} \\
    \midrule
    Learning rate & $5\times10^{-5}$ \\
    Weight decay & 0.05 \\
    Batch size & 16 \\
    Sequence length & 128 \\
    Early stopping patience & 3 \\
    Seed & 12 \\
    
    \bottomrule
    \end{tabular}
    \caption{Training and optimization settings for the main experiments.}
    \label{tab:app:training-hyperparameters}
\end{table}

\section{Evaluation}
\label{sec:app:BLM_evals}

\begin{table*}[htb]
    \centering
    \small
    \begin{tabular}{lllll}
        \toprule
        & \textbf{Dataset} & \textbf{Language} & \textbf{Prediction Task} & \textbf{Reference} \\
        \midrule
        \multirow{7}{*}{\rotatebox{90}{Zero-shot}}
        & \texttt{BLiMP}       & EN         & Grammatical acceptability & \citet{warstadt-etal-2020-blimp} \\
        & \texttt{BLiMP-NL}    & NL         & Grammatical acceptability & \citet{suijkerbuijk-etal-2025-blimpnl} \\
        & \texttt{ZhoBLiMP}    & ZH         & Grammatical acceptability & \citet{liu-etal-2026-systematic} \\
        & \texttt{Hanzi}       & ZH         & Grammatical acceptability & \citet{hu-etal-2026-evaluation} \\
        & \texttt{MultiBLiMP}  & EN, NL     & Grammatical acceptability & \citet{jumelet-etal-2026-multiblimp} \\
        & \texttt{GlobalPIQA}  & EN, NL, ZH & Commonsense reasoning     & \citet{chang-others-2026-global} \\
        & \texttt{MECO}        & EN, NL, ZH & Eye-tracking              & \citet{siegelman-etal-2022-expanding,siegelman-etal-2025-wave,kuperman-etal-2025-new} \\
        & \texttt{HellaSwag}   & EN, NL, ZH & Sentence completion       & \citet{zellers-etal-2019-hellaswag} \\
        & \texttt{Winogrande}  & EN, NL, ZH & Coreference resolution    & \citet{sakaguchi-etal-2021-winogrande} \\
        & \texttt{XStoryCloze} & EN, NL, ZH & Story completion          & \citet{lin-etal-2022-fewshot} \\
        & \texttt{XCOMPS}      & NL, ZH     & Property knowledge        & \citet{he-etal-2025-xcomps} \\
        \midrule
        \multirow{7}{*}{\rotatebox{90}{Fine-tuning}}
        & \texttt{ARC}         & EN, NL, ZH & Abstraction and reasoning & \citet{clark-etal-2018-think} \\
        & \texttt{Belebele}    & EN, NL, ZH & Reading comprehension     & \citet{bandarkar-etal-2024-belebele} \\
        & \texttt{BMLama}      & EN, NL, ZH & Knowledge probing         & \citet{qi-etal-2023-crosslingual} \\
        & \texttt{INCLUDE}     & NL, ZH     & Knowledge probing         & \citet{romanou-etal-2025-include} \\
        & \texttt{MNLI}        & EN, NL, ZH & Inference                 & \citet{williams-etal-2018-broadcoverage} \\
        & \texttt{SIB-200}     & EN, NL, ZH & Topic classification      & \citet{adelani-etal-2024-sib200} \\
        & \texttt{TruthfulQA}  & EN, NL, ZH & Knowledge probing         & \citet{lin-etal-2022-truthfulqa} \\
        & \texttt{POS}        & EN, NL, ZH     & POS-tagging            & \citet{nivre-etal-2020-universal} \\
        & \texttt{XNLI}        & EN, ZH     & Inference                 & \citet{conneau-etal-2018-xnli} \\
        \bottomrule
    \end{tabular}
    \caption{Overview of evaluation datasets used in the BabyLM 2026 multilingual track \cite{choshen-etal-2026-babylm}.
             EN~=~English, NL~=~Dutch, ZH~=~Chinese. All metrics, except \textsc{MECO}, accuracies. Note that \textsc{MECO} is therefore excluded from \cref{fig:blm_eval} and \cref{fig:bayes_model}.} 
    \label{tab:app:evaluation_setup}
\end{table*}

We use the \href{https://github.com/babylm/evaluation-pipeline-2025}{evaluation pipeline} of the 2026 BabyLM Challenge \cite{choshen-etal-2026-babylm}. \Cref{tab:app:evaluation_setup} lists the multilingual evaluation data for zero-shot and fine-tuning tasks.

\section{Metric correlation}
\label{sec:app:metric_correlation}

\begin{figure}
    \centering
    \includegraphics[width=\linewidth]{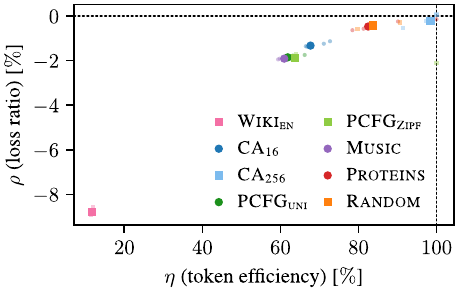}
    \caption{Relation between the two evaluation metrics loss ratio~$\rho$ and token efficiency~$\tau$.}
    \label{fig:app:metric_correlation}
\end{figure}

The two evaluation metrics of loss ratio $\rho$ and token efficiency $\tau$ are strongly related, \cref{fig:app:metric_correlation}, yet loss ratio is more robust to outliers due to early stopping. Token efficiency is highly sensitive to the chosen cutoff point in training steps. Integrating over the loss curve instead, allows for a more robust interpretation of the results.  This robustness is due to the integration of the loss curve, which solves the problem of sensitivity to the cutoff point in training steps that token efficiency suffers from.

\section{Challenge submission}
\label{sec:app:challenge-submission}

For the challenge submission, we train our model on an equal mix of the top three structural data types, namely, on \textsc{Music}, \textsc{CA$_{16}$}, and \textsc{PCFG\textsubscript{Zipf}}. We perform a hyperparameter search with early stopping for the learning rate, weight decay, and warm-up ratio. Next, we select the best configuration and continue parameter search during stage~II. Each stage includes tuning for five seeds and ten epochs. The best model (in terms of validation loss), initialized with the seed 2000, is then evaluated on the zero-shot and fine-tuning tasks of the challenge, and results along with the checkpoints are available on \href{https://hf.co/collections/huds-uni-goe/blm2026}{HuggingFace}. An overview of the searched hyperparameter space is given in the \cref{tab:app:challenge-hyperparameters}; all other parameters are kept constant, as listed in \cref{tab:app:training-hyperparameters}. Model tuning was performed on a single NVIDIA A100 80GB GPU for $\approx 2250$ GPU hours, where each stage-I run was $\approx 4$ hours and each stage-II run was $\approx 15$ hours. The best model along with the results has been uploaded to HuggingFace and BabyLM leaderboard. Results are available in \cref{tab:app:submission-results}.

\begin{table}[htbp]
    \centering
    \small
    \begin{tabular}{ll}
        \toprule
        \textbf{Hyperparameter} & \textbf{Value} \\
        \midrule
        Optimizer learning rate & $1\times10^{-4}$, $3\times10^{-4}$, \\
        & $6\times10^{-4}$, $1\times10^{-2}$ \\
        Weight decay & 0.01, 0.05, 0.1 \\
        Warmup ratio & 0.01, 0.03 \\
        Seeds & 11; 17; 42; 2{,}000; 3{,}407 \\
        \bottomrule
    \end{tabular}
    \caption{Hyperparameter search space for the challenge submission.}
    \label{tab:app:challenge-hyperparameters}
\end{table}

\begin{table}[htbp]
    \centering
    \small
    \begin{tabular}{lccc}
        \toprule
        \textbf{Benchmark} &
        \textbf{English} &
        \textbf{Dutch} &
        \textbf{Chinese} \\
        \midrule
        MultiBLiMP                     & 0.88 & 0.91 & --   \\
        BLiMP                          & 0.72 & 0.78 & 0.76 \\
        WinoGrande                     & 0.52 & 0.49 & 0.49 \\
        XStoryCloze                    & 0.49 & 0.48 & 0.49 \\
        HellaSwag                      & 0.27 & 0.26 & 0.27 \\
        XCOMPS                         & --   & 0.53 & 0.53 \\
        Global PIQA (parallel)         & 0.27 & 0.23 & 0.18 \\
        Global PIQA (non-para.)        & 0.52 & 0.51 & 0.48 \\
        \midrule
        ARC                            & 0.23 & 0.25 & 0.24 \\
        Belebele                       & 0.30 & 0.23 & 0.27 \\
        BMLAMA                         & 0.10 & 0.12 & 0.10 \\
        MNLI                           & 0.58 & 0.59 & 0.57 \\
        SIB-200                        & 0.33 & 0.26 & 0.31 \\
        TruthfulQA                     & 0.25 & 0.26 & 0.29 \\
        XNLI                           & 0.53 & --   & 0.53 \\
        INCLUDE                        & --   & 0.29 & 0.26 \\
        POS                            & 0.89 & 0.91 & 0.81 \\
        \bottomrule
    \end{tabular}
    \caption{Evaluation results for the submission model trained on the mixed dataset in stage~I and on a mix of English, Dutch, and Chinese in stage~II.}
    \label{tab:app:submission-results}
\end{table}

\end{document}